\documentclass{nle}

\usepackage{multirow}
\usepackage[utf8]{inputenc}
\usepackage{booktabs}
\usepackage{xspace}
\usepackage{graphicx}
\usepackage{subfig}
\usepackage{amsmath}
\usepackage{relsize}
\usepackage{covington}

\usepackage{hyperref}
\usepackage{todonotes}

\usepackage[all]{nowidow}

\newcommand\eg{{\it e.g.\ }}

\newcommand\ie{{\it i.e.\ }}
\newcommand\F{{$\text{F}_1$\ }}
\newcommand{\sd}[1]{\par \scriptsize #1}

\definecolor{lightgreen}{RGB}{200,255,200}
\definecolor{lightblue}{RGB}{200,200,255}
\definecolor{lightred}{RGB}{255,200,200}
\newcommand{\posbox}[1]{{\setlength{\fboxsep}{1pt}\colorbox{lightblue}{#1}}}
\newcommand{\negbox}[1]{{\setlength{\fboxsep}{1pt}\colorbox{lightred}{#1}}}

\makeatletter
\def\cite{\def\citename##1{##1 }\@internalcite}
\def\newcite{\def\citename##1{{\frenchspacing##1} (}\@internalciteb}

\def\@citexb[#1]#2{\if@filesw\immediate\write\@auxout{\string\citation{#2}}\fi
  \def\@citea{}\@newcite{\@for\@citeb:=#2\do
    {\@citea\def\@citea{;\penalty\@m\ }\@ifundefined
       {b@\@citeb}{{\bf ?}\@warning
       {Citation `\@citeb' on page \thepage \space undefined}}%
{\csname b@\@citeb\endcsname}}}{#1}}
\def\@internalciteb{\@ifnextchar [{\@tempswatrue\@citexb}{\@tempswafalse\@citexb[]}}

\def\@newcite#1#2{{#1\if@tempswa, #2\fi)}}

\makeatother

\title[Multi-task Learning of Negation for Sentiment]{Improving Sentiment Analysis with \\Multi-task Learning of Negation}
\author[J. Barnes, E. Velldal, and L. Øvrelid]
       {J\ls E\ls R\ls E\ls M\ls Y\ns B\ls A\ls R\ls N\ls E\ls S\ls ,\ns E\ls R\ls I\ls K\ns V\ls E\ls L\ls L\ls D\ls A\ls L\ls ,\ns  and\ns L\ls I\ls L\ls J\ls A\ns Ø\ls V\ls R\ls E\ls L\ls I\ls D \\   
Language Technology Group, University of Oslo \\
email: \{jeremycb,erikve,liljao\}@ifi.uio.no}

\pagerange{\pageref{firstpage}--\pageref{lastpage}}
\pubyear{2019}

\begin{document}

\label{firstpage}
\maketitle

\begin{abstract}

Sentiment analysis is directly affected by compositional phenomena in language that act on the prior polarity of the words and phrases found in the text. \emph{Negation} is the most prevalent of these phenomena and in order to correctly predict sentiment, a classifier must be able to identify negation and disentangle the effect that its scope has on the final polarity of a text. This paper proposes a multi-task approach to explicitly incorporate information about negation in sentiment analysis, which we show outperforms learning negation implicitly in an end-to-end manner. We describe our approach, a cascading and hierarchical neural architecture with selective sharing of LSTM layers, and show that explicitly training the model with negation as an auxiliary task helps improve the main task of sentiment analysis. The effect is demonstrated across several different standard English-language data sets for both tasks and we analyze several aspects of our system related to its performance, varying types and amounts  of input data and different multi-task setups.

\end{abstract}

\section{Introduction}

The goal of sentiment analysis is to assign a polarity (either categorical or real valued) to text and has become a popular task in natural language processing thanks to a growing interest in automatically processing the large amount of opinionated text available on the internet. Consider the following example sentence from a movie review, taken from the Stanford Sentiment Treebank \cite{Socher2013b}, where we have added annotations to indicate words with prior positive polarity (blue boxes), negation cues (bold face), and the scopes of the cues (underlined): 
\begin{examples}
\item\label{ex1}  Being \posbox{unique} \textbf{doesn't} \underline{necessarily equate to being \posbox{good}}, \textbf{no matter} \underline{how \posbox{admirably} the filmmakers have gone for broke}.
\end{examples}
In this short sentence, there is a subtle negative sentiment expressed towards the movie through the negation of the phrase ``necessarily equate to being good''. This example points out how the sentiment of a sentence is not merely the sum of the polarity of the words and phrases found in the text, but rather depends on a number of compositional phenomena that act on indicators of polarity. \textit{Negation} is one of the most pervasive of these phenomena. 

In order to adequately deal with the phenomenon of negation in sentiment analysis, it is not enough to simply detect single words indicating negation, so-called negation cues, as the scope of this negation is equally important. In Example~(\ref{ex2}) below, there is negation, but the relevant polar adjectives ``unique'' and ``well-crafted'' are not within its scope (the red box indicates prior negative polarity).
\begin{examples}
\item\label{ex2} It's \textbf{not} \underline{so much a work of entertainment} as it is a \posbox{unique}, \posbox{well-crafted} psychological study of \negbox{grief}.
\end{examples}
A sentiment classification system that takes a naive view of negation would likely classify the sentence in (\ref{ex2}) as negative, as negation cues often lead models to predict more negative sentiment \cite{Wiegand2010,barnes-etal-2019}. Previous research also demonstrates the need and utility for incorporating negation information in sentiment models \cite{Wiegand2010,councill-etal-2010-great,Lapponi2012,Cruz2016}. Approaches that use negation information to improve sentiment analysis can largely be divided into three broad categories:

\begin{enumerate}
    \item approaches that use \emph{heuristic polarity modification} where the prior polarity of a word is modified if found within some given radius of a negation cue  \cite{HuandLiu2004,Taboada2011},
    \item approaches that \emph{augment the classification feature space} with negation-relevant features \cite{Pang2002,Das2007,Lapponi2012},
    \item or \emph{end-to-end approaches} where the model is assumed to capture the effects of negation without being provided explicit negation annotations \cite{Socher2013b,Irsoy2014}. 
\end{enumerate}

 However, most of the previous approaches to incorporating negation information into sentiment modeling do not take full advantage of the large body of work that exists on negation detection as a task of its own, both in terms of modeling \cite{Mor:Dae:09,ReaVelOvr12b,Fancellu2016} and data sets \cite{Morante2012,Konstantinova2012}. One likely reason for this is that it is not obvious how to best incorporate negation information into state-of-the-art sentiment models. In this paper, we apply multi-task learning to incorporate information from data sets explicitly annotated for negation in order to improve the performance of sentiment classifiers on English-language data sets.  

\paragraph{\textbf{Contributions: }} In this work, we make the following contributions: 

\begin{enumerate}
\item we propose a hierarchical multi-task learning approach to incorporate negation information into a sentiment classifier,
\item we show that multi-task learning can lead to improvements despite a difference in the relevant units of classification (\eg\ sentence-level sentiment and sequence-labeled negation scope),
\item we provide a detailed analysis of the effects of multi-task learning of negation for sentiment analysis.
\end{enumerate}

We additionally make the data and code available\footnote{\url{https://github.com/ltgoslo/multitask_negation_for_sa}} in order to encourage reproducibility. In the remainder of the paper we first discuss related work (Section \ref{sec:related}), then describe the data used in all experiments (Section \ref{sec:data}), and detail our proposed cascading multi-task model in Section \ref{sec:model}. We then describe the results of the main experiment (Section \ref{sec:results}) and perform a thorough analysis of the most important variables in Section \ref{sec:analysis}. Finally, we discuss the implications of our findings and future work in Section \ref{sec:conclusion}.

\section{Related work}
\label{sec:related}

This section first outlines some of the previous work done on handling negation -- both as a part of sentiment analysis and as a separate task in itself. We then review some relevant previous work on sentiment analysis more generally, and finally provide some background on previous work on multi-task learning in NLP. 

\subsection{Negation in sentiment models}
\label{sec:related:sent}

Negation is a frequent linguistic phenomenon which has a direct impact on sentiment analysis \cite{Wiegand2010}. Within the framework of lexicon-based sentiment analysis, researchers first attempted to model negation with simple heuristics, such as reversing \cite{HuandLiu2004,Polanyi2006,Kennedy2005} or modifying \cite{Taboada2011} the polarity signal of a negated word. This approach to tackle contextual valence shifting generally assumes that the final polarity of a text is some function of the prior polarities of adjectives, verbs, and nouns found in the text. The scope of negation is determined heuristically, by finding common negation cues and assuming all words between the cue and the next punctuation are in scope \cite{HuandLiu2004} or based on the distance from the cue \cite{Taboada2011}.

Early machine learning approaches to sentiment analysis also used heuristics, such as attaching a negation tag (``\_neg'') to words assumed to be in scope \cite{Pang2002,Das2007}. This approach, however, leads to an increase in sparsity and varying results, as the sentiment model is not able to explicitly connect the original and negated features. Other research has used negation detection systems to enhance the feature space of sentiment models \cite{councill-etal-2010-great,Lapponi2012,Cruz2016}, leading to improved results on sentiment classification. Additionally, certain negation cues contribute to higher shifts in polarity than others \cite{Zhu2014}, which indicates they should be modelled separately.

%\newcite{Zhu2014} perform a detailed quantitative study on the effect of negation words on sentiment. They modify the Recursive Neural Tensor Network model \cite{Socher2013b} to more directly model the relationship between negation cues and polar adjectives, which gives a small boost in performance. They conclude that certain negation cues contribute to higher shifts in polarity than others and therefore should be modelled separately.

More recent advances in negation detection, both in terms of modeling and data annotation, have not been incorporated into sentiment classification models so far, to the best of our knowledge.

\subsection{Negation detection}
\label{sec:related:neg}

\begin{figure}[]
\begin{center}
\includegraphics[width=\textwidth]{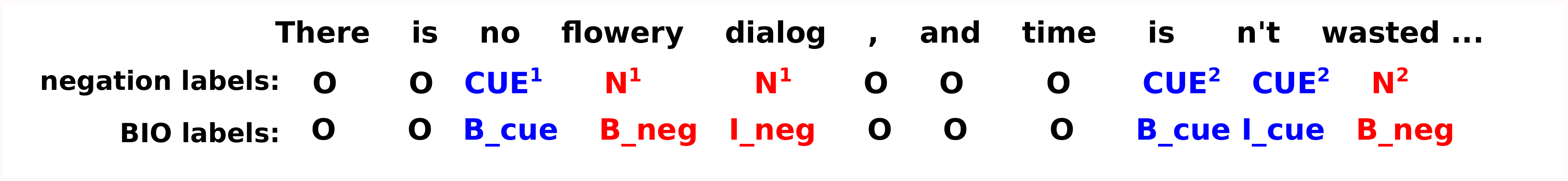}
\caption[]{An example of negation detection annotation on a sample sentence from the SFU dataset.}
\label{fig:neg-labels}
\end{center}    
\end{figure}

Previously reported approaches to negation analysis commonly breaks it down to (at least) two sub-tasks, performing (i) negation cue detection, followed by (ii) scope detection. The example in Figure \ref{fig:neg-labels} shows the negation annotation of the following sentence (in the first row):
\begin{examples}
\item\label{fig1} There is  \textbf{no} \underline{flowery dialog}, and time \textbf{isn't} \underline{wasted}.
\end{examples}

For our example in Figure \ref{fig:neg-labels} the cue detection component would locate the negation cues in the sentence, i.e., the negative determiner ``no'', and the copula with its negative contraction ``is n't'', whereas the scope detection module would recognize the noun phrase ``flowery dialog'' and the verb phrase ``wasted'' as the scopes of these cues, respectively. Depending on the specific annotation scheme, subjects may or may not be part of the scope of negation. 

A large portion of early work on negation detection  \cite{Mor:Lie:Dae:08,Mor:Dae:09,VelOvrRea12} has been done within the biomedical domain due to the availability of the BioScope corpus \cite{Vin:Sza:Far:08}, which is annotated for negation cues and scopes. 
Interest in the task was further spurred by the *SEM shared task \cite{Morante2012}, which focused on detection of negation cues and scopes, in addition to detection of negated events and their so-called focus. The shared task made available the ConanDoyle-neg corpus, which is described in Section~\ref{sec:data} below. A number of systems were submitted for this task, employing a wide variety of strategies. For example, the best performing systems for the closed track and open track employed, respectively, SVM-based ranking of constituent (sub-)trees \cite{ReaVelOvr12b} and CRF-based sequence-labeling using dependency features \cite{LapVelOvr12}. 

Traditional approaches to the task of negation detection have typically employed a wide range of hand-crafted features describing a number of both lexical, morphosyntactic and even semantic properties of the input text. Syntactic parsing has often been used to analyze the input prior to negation detection and has been based on both constituency-based \cite{ReaVelOvr12b,Pac:Ben:Rea:2014} and dependency-based representations \cite{Lapponi2012,Whi:2012,EngVelOvr17}. It is also possible to combine an existing system \cite{ReaVelOvr12b} with an additional layer of manually defined rules over Minimal Recursion Semantics structures created by an HPSG parser \cite{Pac:Ben:Rea:2014}.

There are a few previous studies that investigate neural modeling for the task of negation detection. Among these we find a CNN model for the negation scope detection on the abstracts section of the BioScope corpus, which operates over syntactic paths between the cue and candidate tokens \cite{Qia:Li:Zhu:2016}. \cite{Fancellu2016} present and compare two neural architectures for the task of negation scope detection on the ConanDoyle-neg corpus: a simple feed-forward network and a bidirectional LSTM. 
%The approach uses only word embeddings as the input representation, dispensing with explicit syntactic features. 
Note that these more recent neural systems disregard the task of cue detection altogether  \cite{Fancellu2016,Qia:Li:Zhu:2016,fancellu-etal-2017-detecting}, relying instead on gold cues and focusing solely on the task of scope detection.

While syntactic information has often been found useful for scope resolution, the task of cue detection appears to only require simpler surface information. \cite{VelOvrRea12} present an approach to cue detection which treats the set of cue words as a closed class and apply a disambiguation-based approach to the problem of cue detection, showing that simple lexical features based on a narrow context window is sufficient to achieve good performance. 

% include after the survey of previous work on negation detection: 
As further detailed in Section~\ref{sec:model:negation}, in the current paper we model cue detection and scope resolution concurrently as a sequence-labeling task, using a BIO label encoding which is illustrated in the final row of Figure \ref{fig:neg-labels}. BIO-labeling for negation detection has been employed in previous work, following the early work of \cite{Mor:Lie:Dae:08}. Our joint modeling of cue and scope detection differs from previous approaches to negation detection as reviewed above that handle cue and scope resolution as two separate tasks \cite{Morante2012,LapVelOvr12,ReaVelOvr12b,Fancellu2016,Cruz2016,Qia:Li:Zhu:2016}. Moreover, motivated by the assumption that downstream tasks like sentiment analysis only need information about which words are within the scope of some negation, regardless of which particular cue it relates to (in cases where more than one cue is present), we do not attempt to explicitly retain this coupling. Similarly to \cite{Fancellu2016} we use a BiLSTM-based model, relying only on word embeddings as input, but also adding a CRF for the prediction layer. 
When it comes to incorporating information about negation to our sentiment model, our approach takes advantage of the representation learning capabilities of neural models: rather than passing on the negation predictions output by the final CRF layer, we pass on the intermediate representations learned by the BiLSTM. The details of this cascading architecture are further described in Section~\ref{sec:model}.

\subsection{Sentiment analysis}
\label{sec:related:only-sent}

Approaches to sentiment analysis have moved from lexicon-based methods \cite{Turney2002,HuandLiu2004,Taboada2011}, to machine learning methods based on hand derived features \cite{Pang2002,PangLee2008} and finally to neural networks that learn to extract useful features in an end-to-end fashion \cite{Socher2013b,Tang2014,Tai2015a}. While some of these neural architectures have been tailored to suit specific tasks better \cite{Irsoy2014,Lei2018}, two recent end-to-end architectures have shown competitive results on a large number of natural language processing tasks: bidirectional Long Short-term Memory Networks (BiLSTMs) \cite{Graves2005} and Self-Attention Networks (SANs) \cite{Vaswani2017}. Variants of these two architectures give state-of-the-art results on document-level \cite{howard-ruder-2018-universal},  sentence-level \cite{Peters2018,Devlin2018}, and aspect-level \cite{Xu2019} sentiment analysis tasks.

The claim made by the proponents of end-to-end learning is that the models implicitly learn compositional functions \cite{Socher2013b,Irsoy2014}, thereby removing the need to explicitly provide information about inter-word dependencies, negation, or speculation in the form of hand-crafted features. Recent research, however, challenges the idea that end-to-end learning is able to fully capture compositional effects \cite{verma-etal-2018-syntactical,barnes-etal-2019}. It is therefore worth asking whether we can help the model by providing some form of explicit training on compositional phenomena in sentiment.

\subsection{Multi-task learning}
\label{sec:related:mtl}

Multi-task learning (MTL) \cite{Caruana93multitasklearning} stems from the idea that learning related tasks simultaneously allows a machine learning algorithm to incorporate a useful inductive bias by restricting the search space of possible representations to those that are predictive for both tasks. MTL assumes that features that are useful for a certain task should also be predictive for similar tasks, and in this sense MTL also effectively acts as a regularizer, as it prevents the weights from adapting too much to just one task. Under some circumstances, multi-task learning can also be seen as a kind of data augmentation, where an MTL model takes advantage of extra training data available in an auxiliary task to improve the main task \cite{kshirsagar-etal-2015-frame,plank-2016-keystroke,FarOepVel18}. MTL is particularly well-suited for neural models, given the possibilities for modular design and representation learning. Below we outline some of the different ways that multi-task learning can be set up, while also reviewing previous MTL efforts in NLP.

\textit{Hard parameter sharing} \cite{Caruana93multitasklearning}, which assumes that all layers are shared between tasks except for the final predictive layer, is the simplest way to implement a multi-task model. When the main task and auxiliary task are closely related, this approach has been shown to be an effective way to improve model performance \cite{Collobert2011a,peng-dredze-2017-multi,Alonso2017,Augenstein2018}. Other research \cite{Sogaard2016}, on the other hand, finds that it is better to make predictions for low-level auxiliary tasks at lower layers of a multi-layer MTL setup. They also suggest that under the hard-parameter framework auxiliary tasks need to be sufficiently similar to the main task for MTL to improve over the single-task baseline.

There have also been several effective implementations of \textit{soft parameter sharing}, where two models have both shared and private task-specific parameters, such as including a gating mechanism that allows a MTL model to select which information to share across tasks \cite{Liu2016,Ruder2019}. Their results suggest that hard parameter sharing is only beneficial for low-level tasks, while for high-level tasks it is better to learn how much to share at each layer and subspace of parameters in the network. Furthermore, they find that MTL is more beneficial when there is less training data for the main task and that modeling subspaces explicitly helps in almost all domains.

What characteristics of an auxiliary task are necessary to improve a main task is still largely unknown. Some research suggests high-level semantic auxiliary tasks generally help more than low-level auxiliary tasks as MTL tends to work when the main task learning plateaus quickly and the auxiliary task learning does not \cite{Bingel2017}. Others find that auxiliary tasks with compact, uniform label distributions are preferable \cite{Alonso2017}. Additionally, choosing a suitable auxiliary task is still vital, as introducing an unsuitable auxiliary task can actually hurt performance \cite{Augenstein2017}. %\newcite{Augenstein2017} use MTL to improve key-phrase boundary classification in scientific articles. They use hard parameter sharing with five auxiliary tasks (chunking, frame prediction, hyperlink prediction, multi-word identification, and super-sense tagging) and find that their models outperform the previous state-of-the-art. Choosing the correct auxiliary task, however, depends on the target task, as introducing an unsuitable auxiliary task can actually hurt performance.

%Building on this last observation, \newcite{Alonso2017} explore which kinds of low-level auxiliary tasks improve performance in a MTL setup for semantic sequence-prediction tasks, finding that auxiliary tasks with compact, uniform label distributions (such as POS tagging or frequency bin prediction) are preferable, but that only 1 in 5 MTL setups contribute to improved results. They conclude, however, that the problem may lie more in the skewness of the data than the difficulty of MTL for semantic tasks.

In this work, we propose that MTL is an appropriate framework to incorporate negation detection in a sentiment classifier. Unlike previous approaches in sentiment analysis \cite{councill-etal-2010-great,Lapponi2012,Cruz2016}, 
our method does not rely on incorporating negation information as explicit features, but rather uses a cascading architecture where the intermediate representations learned for predicting negation feeds into subsequent layers (along with skip-connections) for predicting sentiment. While the final layers of the network hierarchy are dedicated to the sentiment task, the lower layers are shared and supervised by both tasks. The components of the architecture are further detailed in Section~\ref{sec:model}. Note that, for comparison, we also explore using other auxiliary tasks beyond negation. 

In parallel work to this, \cite{BarSamOvr19} showed that a similar cascading or hierarchical MTL architecture could be used for incorporating information from sentiment lexicons to improve models for sentence-level sentiment classification. Also in parallel work, \cite{SanWolRud19} apply hierarchical MTL to learn shared representations for a set of semantic tasks where lower-level task like named entity recognition and entity mention feeds into higher-level tasks like coreference resolution and relation extraction.

\section{Data sets}
\label{sec:data}

As outlined above, we propose to model both sentiment classification and negation detection in a multi-task learning set-up. Unlike much previous work in MTL \cite{Bingel2017,Augenstein2017,Bjerva2017,Ruder2019} which assumes several prediction tasks annotated on the same dataset with the same output units (token-level sequence labeling), we take auxiliary data from different data sets and domains and with different units of classification across tasks: We experiment with sentence- and tweet-level classification of sentiment as a main task, while learning sequence-labeling of negation cues and scopes based on two different negation data sets as an auxiliary task. This section describes the different data sets we use.

\paragraph{SFU Review Corpus:} This corpus \cite{Konstantinova2012} contains 400 reviews from eight domains (books, cars, computers, cookware, hotels, movies, music, phones) which have been annotated for sentiment at document-level, as well as negation and speculation at sentence-level. Although the dataset contains sentiment annotations, we do not use these to evaluate the sentiment models, but rather choose to focus on sentence- and tweet-level classification, as compositional effects will have a more direct bearing on the prediction on these finer-grained tasks. The example in Figure \ref{fig:neg-labels} illustrates the annotation scheme found in the SFU corpus (top rows). The annotation scheme is based principally on the guidelines developed for the biomedical BioScope corpus \cite{Vin:Sza:Far:08}, which largely employ syntactic criteria for the determination of negation scope, choosing the maximal syntactic unit that contains the negated content. Unlike BioScope, however, negation cues are not included within the scope. The SFU corpus does not annotate affixal cues, e.g. \textit{im-} in \textit{impossible}. This corpus, however, has the advantage that it stems from the same domain (reviews) as our main task. As there is not a predefined test split, we take 800 sentences annotated for negation as training, 71 for development, and 96 for testing.

\paragraph{ConanDoyle-neg (CD):} This widely used corpus contains Conan Doyle stories manually annotated for negation cues, scopes, and events \cite{Mor:Dae:12} and was employed in the 2012 *SEM shared task on negation detection \cite{Morante2012}. The shared task version of the dataset contains a training set of 3,640 sentences, of which 848 sentences contain negation, a development set consisting of 787 sentences, of which 144 are negated, as well as a held-out test set which was constructed specifically for the shared task, consisting of 1089 additional sentences, of which 235 sentences contain negation. The annotation scheme is also based on those employed for the BioScope corpus \cite{Vin:Sza:Far:08}, but with some important modifications. In ConanDoyle-neg (CD hereafter), the cue is not included in the scope, and it annotates a wide range of cue types, i.e., both sub-token (affixal), word-based and multi-word negation cues. Scopes may furthermore be discontinuous, often an effect of the requirement to include the subject within the negation scope. This is in contrast to the annotation scheme found in the SFU corpus, where subjects are not included in negation scope, as is clear from the example in Figure \ref{fig:neg-labels}, where the subject \textit{time} is not included in the scope of the negation cue \textit{isn't}.
In our experiments, we use the pre-defined train, development, and test splits from the shared task.

\paragraph{Stanford Sentiment Treebank (SST):} The SST data \cite{Socher2013b} contains 11,855 sentences taken from English-language movie reviews. It was annotated for fine-grained sentiment (strong negative, negative, neutral, positive, strong positive) which we refer to as SST-fine and can also be mapped to a binary setting (SST-binary), where the neutral class is removed and strong and normal examples are merged (9,613 sentences). We perform experiments with both setups, using the pre-defined train, development and test splits.

\paragraph{SemEval 2013:} The SemEval 2013 shared task on tweet-level sentiment analysis \cite{Nakov2013} contains 9,287 tweets annotated for three-way sentiment (positive, neutral, negative), which we refer to as SemEval-fine. Additionally, we remove the tweets with neutral labels to give a binary setup (SemEval-binary). We use the train, development and test splits given in the shared task.

\begin{table}
\caption{Overview of the data sets for sentiment and negation. Note that for SST, SFU, and Conan Doyle Neg, we show the number of sentences, while for SemEval 2013 we show the number of tweets.}
\begin{minipage}{\textwidth}
\footnotesize
\begin{tabular}{cllrrr}
\hline\hline
Task & & Dataset & \multicolumn{1}{c}{Train} & \multicolumn{1}{c}{Dev.} & \multicolumn{1}{c}{Test} \\
\hline
\multirow{4}{*}{Sentiment} & \multirow{2}{*}{\cite{Socher2013b}} & SST-binary & 6,920 & 872 & 1,821 \\
&& SST-fine & 8,455 & 1,101 & 2,210 \\
\cmidrule(){2-2}\cmidrule(){3-3}\cmidrule(){4-4}\cmidrule(){5-5}\cmidrule(){6-6}
& \multirow{2}{*}{\cite{Nakov2013}} &SemEval2013-binary &  3,056 & 491 & 1,262 \\
&& SemEval2013-fine &    6,021 & 890 & 2,376\\
\cmidrule(){2-2}\cmidrule(){3-3}\cmidrule(){4-4}\cmidrule(){5-5}\cmidrule(){6-6}
\multirow{2}{*}{Negation} & \cite{Konstantinova2012} & SFU Review Corpus &  800 & 71 & 96 \\
\cmidrule(){2-2}\cmidrule(){3-3}\cmidrule(){4-4}\cmidrule(){5-5}\cmidrule(){6-6}
& \cite{Morante2012} & Conan Doyle Neg & 842 & 144 & 235 \\
\hline\hline
%\bottomrule
\end{tabular}
\end{minipage}
\end{table}

\section{A cascading multi-task model for negation and sentiment}
\label{sec:model}

\begin{figure}
\includegraphics[width=\textwidth]{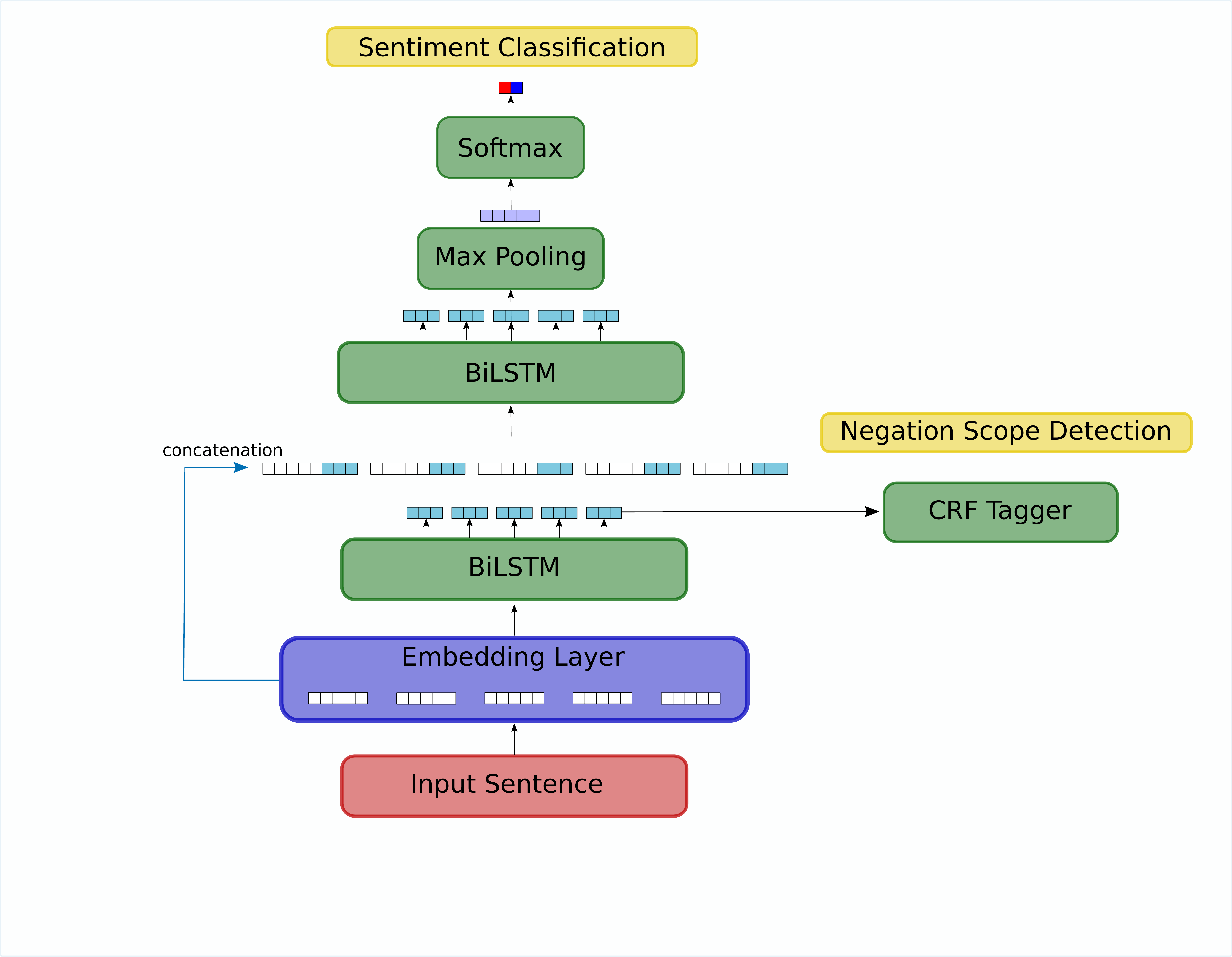}
\caption[Multi-task model]{Our proposed multi-task model.}
\label{figure:model}
\end{figure}

This section details our neural architecture for multi-task learning of negation and sentiment, as shown in Figure~\ref{figure:model}. We adopt a cascading architecture where the lower layers are used to perform the auxiliary task -- in our case negation cue and scope prediction -- and the higher layers are dedicated to the main task -- in our case polarity prediction. Adopting the terminology of \cite{Goldberg:17}, `cascading' here refers to the fact that rather than passing on the negation predictions as such, the lower layers passes on the intermediate representations learned for making these predictions. The multi-task learning set-up means that the shared lower layers will receive supervision signals from both the sentiment and negation tasks. This set-up also aligns well with the findings of \cite{Sogaard2016} that MTL models tend to benefit more from lower-level auxiliary tasks at lower layers of the network. We detail the different components in more detail below.

\subsection{Negation model}
\label{sec:model:negation}

We start by discussing the part of the model responsible for detecting negation cues and scopes, including how it relates to some of the previously reported approaches that are most directly relevant.

Similarly to \cite{Fancellu2016}, we use a bidirectional Long Short-term Memory (BiLSTM) network \cite{Graves2005} to extract features from the embedding layer,  but where \cite{Fancellu2016} use a linear softmax layer for prediction, we use a linear-chain conditional random field (CRF) with Viterbi decoding to find the most probable assignment of labels. Moreover, while \cite{Fancellu2016} assume gold cues, encoded as separate cue embeddings concatenated to the word embeddings provided as input, we here let the BiLSTM predict both cues and scopes -- performed in one pass.

Note that there might be several instances of negation in the same sentence, as in the example of Figure~\ref{fig:neg-labels}. In the set-up of \cite{Fancellu2016}, each instance is multiplied out into a separate example, effectively duplicating the sentence for each pair of cue and scope. In our set-up, all instances will be treated in the same pass. In the CRF model of \cite{LapVelOvr12} too, all scopes are predicted in one pass -- although cues are there predicted in a preceding step using an SVM classifier as in \cite{ReaVelOvr12b} -- but then post-processing heuristics are applied to assign the identified negation tokens to their respective cues. 

A simplifying assumption made in our model is that we do not care about explicitly preserving the links between particular cues and scopes in our output; intuitively, the important information for a downstream task like sentiment analysis is whether a token is within the scope of negation, regardless of the identity of the negation cue. Additionally, since we do not incorporate sub-token information in our model, we treat any token annotated with morphological negation, \eg \textit{un-} or \textit{-less}, as a negation cue.

%Given an input sequence of embeddings $X \in \{x_{1}, x_{2}, \ldots, x_{n}\}$, the representation $V$ from a bidirectional LSTM is the concatenation ($\circ$) of outputs from two LSTM models which take the input sequence from different directions (left to right, right to left):
%
%\begin{equation}
%V = \textrm{BiLSTM}(x_{1:n}) = \textrm{LSTM}_{left}(x_{1:n}) \circ \textrm{LSTM}_{right}(x_{n:1})
%\end{equation}

As shown in Figure \ref{figure:model}, given a sequence of tokens, our negation model first embeds these in an embedding layer, then uses a BiLSTM to create a contextualized representation of each token. This representation is then used as features in the CRF. In our experiments, we use Viterbi decoding to find the most probable assignment of labels, and train the model to minimize the negative log likelihood.

%\begin{equation}
%    P(Y|V) = \frac{\mathlarger{\mathlarger{\mathlarger{\Pi}}}_{i=1}^{|V|} \Psi(V, Y_{i}) %\mathlarger{\mathlarger{\mathlarger{\Pi}}}_{j=1}^{|V|} \Phi(V, Y_{i}, Y_{i-1})}{Z(V)}
%\end{equation}
%
%In this definition of a CRF, $\Psi(V,Y_{i})$ represents the output clique potential of $Y_{i}$
%
%\begin{equation}
%\Psi(V,Y_{i})  = exp\{\theta \cdot V \}
%\end{equation}
%
%where $\theta$ is a model parameter.  $\Phi$ represents the transition likelihood where $\tau(Y_{i-1},Y_{i})$ is the transition weight, which is another parameter of the model.
%
%\begin{equation}
%\Phi(V, Y_{i}, Y_{i-1})  = exp\{\tau(Y_{i-1}, Y_{i})\}
%\end{equation}
%
%Finally, $Z$ is a partition function which assures that the distribution sums to one.
%
%\begin{equation}
%    Z(V)  = \sum_{Y^{'}}  \bigg(  \mathlarger{\mathlarger{\mathlarger{\Pi}}}_{i=1}^{|V|} \Psi(V, Y_{i}^{'}) \mathlarger{\mathlarger{\mathlarger{\Pi}}}_{j=1}^{|V|} \Phi(V, Y_{i}^{'}, Y_{i-1}^{'})    \bigg)
%\end{equation}
%
%

\subsection{Sentiment model}
\label{sec:model:sentiment}

The sentiment model uses the same embedding layer and the first BiLSTM layer to create the contextualized representation of the input tokens. We make use of skip-connections where we concatenate each of the original embeddings to the contextualized representations. This sequence then serves as input to a second sentiment-specific BiLSTM layer.

%$H$ is the output of a second BiLSTM layer:
%
%\begin{equation}
%    H = \textrm{BiLSTM}(\{X_{1} \circ V_{1}, X_{2} \circ V_{2}, \ldots, X_{n} \circ V_{n} \})
%\end{equation}

Finally, we perform a max pooling operation on the output of the sentiment-specific BiLSTM and pass this max-pooled representation to a softmax layer to compute the class probabilities. 
% <-- fixme: should we have a reference to Alexis Conneau here?
We then minimize the cross entropy loss of the sentiment predictions with respect to the true sentiment.

During training, the model alternates between training one epoch on the main task and one epoch on the auxiliary task. Preliminary experiments showed that more complicated training strategies (alternating training between each batch or uniformly sampling batches from the two tasks) did not lead to improvements. Note that we do not upsample negation data. We train the model for 10 epochs using Adam \cite{Kingma2014a}, performing early stopping determined by accuracy on the development set. We regularize\footnote{These optimal values were chosen by observing performance on the development set when training only on the main task and kept stable through all experiments.} with dropout before the BiLSTM layers ($0.5$), between BiLSTM layers  ($0.3$), apply batch norm, and L2 regularization ($0.0001$). As neural models are sensitive to the random initialization of their parameters, we perform five runs with different random seeds and show the mean and standard deviation as the final result for each model.\footnote{We use the same five random seeds for all experiments to ensure a fair comparison between models.}

\section{Results and comparison with state-of-the-art}
\label{sec:results}

Table \ref{table:results} shows the mean accuracy and standard deviation of single-task sentiment models (\textbf{STL}), multi-task models with SFU auxiliary negation data (\textbf{MTL-SFU}) and multi-task models with ConanDoyle-neg auxiliary negation data (\textbf{MTL-CD}) over five runs. One important design decision in these experiments is that, in order to isolate the effects of multi-task learning, we make sure all models have the same capacity in terms of number of parameters: The single-task models also include the lower BiLSTM layers, the difference being that they are supervised by the sentiment task only.

It is important to note that the objective of the current paper is not to achieve new state-of-art results for sentiment analysis, but rather to gauge the relative contribution of negation as an auxiliary task using MTL. Nonetheless, we also include a comparison with the following sentiment models:
\begin{itemize}
    \item \textbf{BOW}:  a L2-regularized logistic regression model trained on a bag-of-words representation \cite{Barnes2017}.
    \item \textbf{CNN}:  a one-layer convolutional neural network with one convolutional layer on top of pre-trained word embeddings \cite{Barnes2017}.
    \item \textbf{BiLSTM}:  a bidirectional LSTM creates a hidden representation from pre-trained word embeddings, which is then mean pooled and fed to a feed-forward network \cite{Barnes2017}.
    \item \textbf{SAN+RPR}: a self-attention network with relative postion representations \cite{Ambartsoumian2018}.
    \item \textbf{Tree-LSTM}: a recursive LSTM that uses parse-trees annotated for sentiment at each node as input \cite{Tai2015a}.
    \item \textbf{BERT}: a large self-attention network pre-trained on a cloze-like language modeling task, and then fine-tuned on the main task \cite{Devlin2018}.
    \item \textbf{HEUR}: this model is identical to the STL model, but incorporates a negation embedding, which is learned during training, and is concatenated to the word embeddings before being passed to the LSTM modules. The negation information comes from performing a heuristic negation processing where any token from a negation cue to the next punctuation mark is considered in scope.
\end{itemize}

The single-task model (STL) achieves an average accuracy of 84.57 on SST-binary, 46.49 on SST-fine, 84.0 on SemEval-binary and 67.26 on SemEval-fine. These results are better than standard performance for a Bidirectional LSTM model (82.6/45.6/-/65.1) and competitive with similar models. The improvement most likely derives from the extra BiLSTM layer, skip-connections, and the max-pooling operation before the softmax layer. Previous state-of-the-art BiLSTM models \cite{Barnes2017} instead use a single layer BiLSTM with mean-pooling. The STL model outperforms the SAN+RPR model on SST-binary, but performs worse on the SST-fine and SemEval-fine tasks. The Tree-LSTM outperforms the STL model on the SST data sets while the BERT model is the best performing system overall. Note that these final two approaches have access to a much larger quantity of data than the others, either in the form of phrase-level annotations for the Tree-LSTM or language model pretraining on more than three billion words in the case of BERT. The HEUR method, however, performs poorly, not even reaching the performance of STL.

The MTL models outperfom the STL models on six of the eight experiments. The MTL-SFU model achieves accuracies of 86.04 (+1.47 percentage points (ppt.)), 46.75 (+0.26 ppt.), 84.02 (+0.02 ppt.), and 67.03 (-0.23 ppt.), improving over the STL on the first three tasks, while the MTL-CD model data has an accuracy of 85.43 (+0.86 ppt.), 47.33 (+0.84 ppt.), 83.53 (-0.48 ppt.), and 67.75 (+0.48 ppt.). Interestingly, the MTL-SFU model is the best performing model on both binary tasks, while the MTL-CD model is the best on both fine-grained tasks. Note that while SAN, Tree-LSTM, and BERT perform better than our proposed models, we choose to use BiLSTMs as it is both easier and faster to perform multi-task learning, allowing for a deeper analysis. Given the results, however, it is clear that follow-up work should concentrate on incorporating the negation information into more advanced models.

We test the significance by performing approximate randomization testing \cite{Yeh2000} with 10,000 iterations pairwise between the results of each of the five runs.\footnote{We use a reimplimentation of the \textsc{sigf} package \cite{Sigf06}.} We consider results significant if the difference between models in at least three of the five runs\footnote{Although t-tests are common in such situations, we opted against this as the independence assumptions do not hold.} are statistically significant ($p < 0.01$ which corresponds to a Bonferroni correction for five hypotheses). MTL models perform significantly better than the STL baseline in four of eight experiments.

\begin{table}
\caption{Mean accuracy and standard deviation of STL and MTL models over five runs on the main sentiment task. The MTL model trained with negation outperforms the single-task baseline in both fine-grained and binary setups. \underline{Underlined} results indicate the best overall approach, while \textbf{bold} results show where the MTL model outperforms the STL. A star (*) indicates that the model performs significantly better ($p < 0.01$), according to approximate randomization tests. We do not report results for SemEval-binary for baseline models or SemEval-fine for Tree-LSTM because the previous work does not report results on this data.}
\begin{minipage}{\textwidth}
\footnotesize
\begin{tabular}{lllll}
\hline\hline
 Model & SST-binary & SST-fine & SemEval-binary & SemEval-fine\\
\hline
BOW & 80.7 & 40.3 & -- & 65.5 \\
CNN & 81.3 \sd{(1.1)} & 39.8 \sd{(0.7)} & -- & 63.5 \sd{(1.3)} \\
BiLSTM & 82.6 \sd{(0.7)} & 45.6 \sd{(0.7)} & -- & 65.1 \sd{(0.9)} \\
SAN + RPR & 84.2 \sd{(0.4)} & 48.1 \sd{(0.4)} & -- & 72.2 \sd{(0.8)} \\
Tree-LSTM & 88.0 \sd{(0.3)} & 51.0 \sd{(0.5)} & -- & -- \\
BERT & \underline{94.9} & \underline{53.0} & -- & \underline{75.1} \\
\hline
HEUR &  83.78 \sd{(0.3)} &  45.29 \sd{(1.6)} & 82.59 \sd{(1.4)} & 65.59 \sd{(1.3)} \\
STL & 84.57 \sd{(1.0)} &  46.49 \sd{(0.7)} & 84.00 \sd{(1.0)} & 67.26 \sd{(0.6)} \\
MTL-SFU & \textbf{86.04 \sd{(0.3)}}* & \textbf{46.75 \sd{(0.8)}} & \textbf{84.02 \sd{(0.9)}} & 67.03 \sd{(1.0)} \\
MTL-CD & \textbf{85.43 \sd{(0.9)}}* & \textbf{47.33 \sd{(0.6)}}*  & 83.52 \sd{(1.6)} & \textbf{67.75 \sd{(1.0)}}*\\
\hline\hline
%\bottomrule
\end{tabular}
\end{minipage}
\label{table:results}
\end{table}

\section{Model analysis}
\label{sec:analysis}

In this section, we include detailed analyses of several aspects of our model. The first analysis is an error analysis that gives a more qualitative view of the results (Section \ref{sec:erroranalysis}). We then perform an analysis of the impact of dataset size (both for the main and auxiliary tasks), and dataset composition (Sections \ref{sec:negsize}--\ref{sec:phrase-level}). Finally, we evaluate several components of the multi-task learning setup (Sections \ref{sec:transfer}--\ref{sec:otherauxtasks}).

\subsection{Error analysis}
\label{sec:erroranalysis}

\begin{figure}
    \centering
    \includegraphics[width=\textwidth]{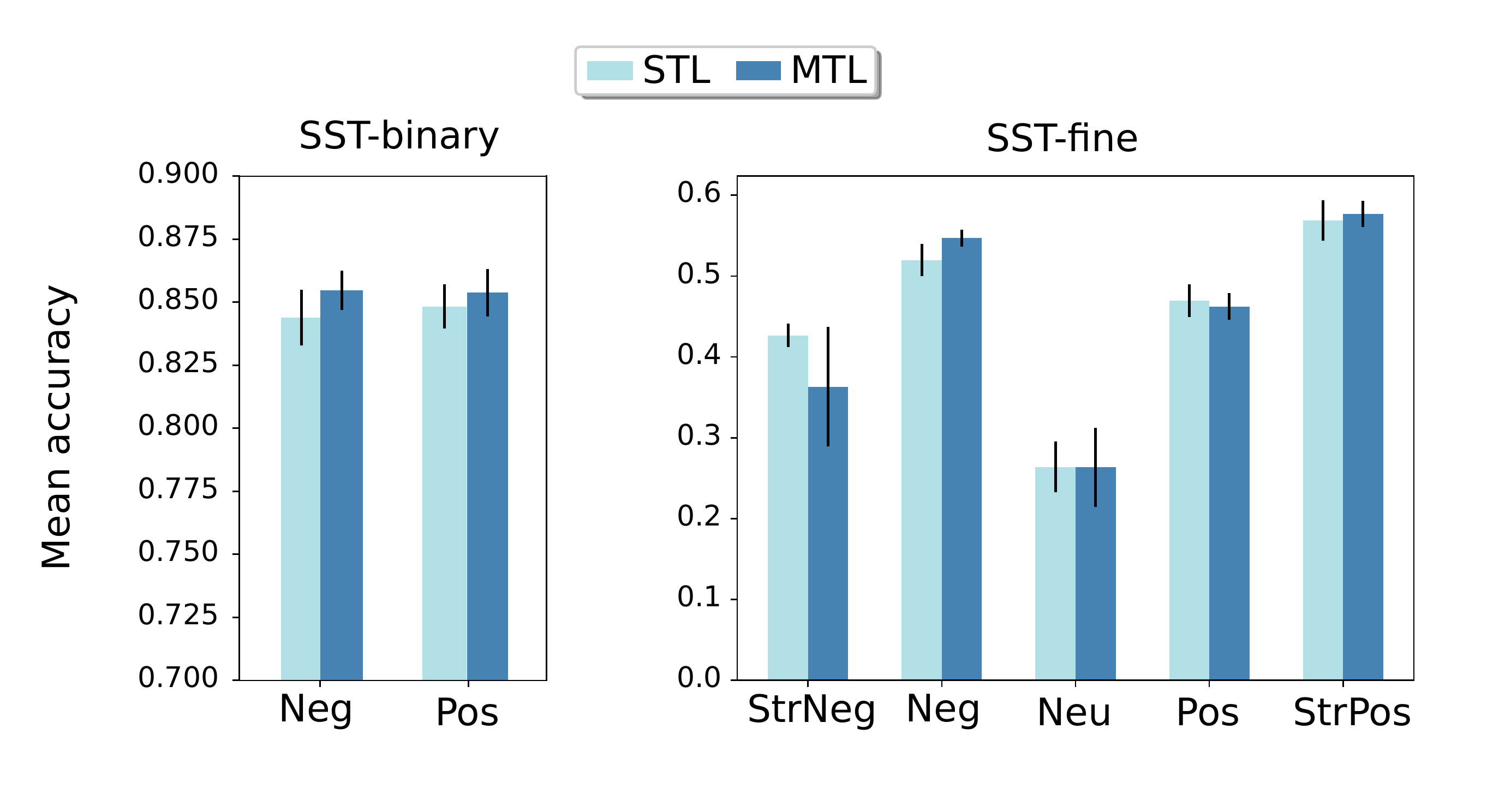}
    \caption{Mean accuracy and standard deviation of STL and MTL-CD model on the SST-binary and SST-fine tasks, broken down across the two (five) classes.}
    \label{figure:perclass}
\end{figure}

A per class evaluation of the SST-binary and SST-fine tasks (Figure \ref{figure:perclass}) shows what effect multi-task learning has on each sentiment class.\footnote{We show both results from the MTL-CD model, in order to isolate the effects of the multitask training from differences in data.} On SST-binary, the MTL model improves on both positive and negative classes. On the SST-fine task, however, the model improves only on the negative and strong positive classes, performing worse on strong negative and positive, while performing nearly the same on neutral. An analysis of the data shows that the negative class contains the largest percentage of negated sentences (27\%), while the strong positive has the least (13\%). It is possible that the MTL model is able to better discriminate relevant and non-relevant negation. In the example from the SST-fine task (\ref{negated_example}) below, the STL model assigned the sentence a positive label due largely to the number of positive tokens, while the MTL-CD model correctly predicted the negative label, as it was able to resolve the negation. 

\begin{examples}
\item\label{negated_example} Accuracy and realism are terrific, but if your film becomes boring, and your dialogue isn't smart, then you need to use more poetic license.
\end{examples}

%\begin{figure}
%\includegraphics[width=\textwidth]{per_class_binary.pdf}
%\caption[]{Mean accuracy and standard deviation of the STL and MTL-CD models on the SST-binary task, broken down across the positive and negative classes.}
%\label{figure:perclassbinary}
%\end{figure}

%\begin{figure}
%\includegraphics[width=\textwidth]{per_class_eval.pdf}
%\caption[]{Mean accuracy and standard deviation of STL and MTL-CD model on the SST-fine task, broken down across the five classes (strong negative/positive, positive/negative and neutral).}
%\label{figure:perclass}
%\end{figure}

The previous analysis suggests that the multi-task setup is beneficial for sentiment analysis, but does not confirm that the model is actually learning better representations for negated sentences. Here, we look at how each model performs on negated and non-negated sentences.

As we do not have access to gold negation annotations on the main task sentiment data, we create silver data by assuming that any sentence that has a negation cue (taken from SFU) is negated. We then extract the negated and non-negated sentences from the SST fine-grained (397 negated / 1813 non-negated) and binary (319 / 1502) test sets. While this inevitably introduces some noise, it allows us to observe general trends regarding these two classes of sentences.

Table \ref{table:analysis} shows the results of the analysis. On the binary task, the MTL model performs better on both the negated (+2.2 ppt.) and non-negated (+1.3 ppt.) subsets. On the fine-grained task, however, the STL model outperforms the MTL model on the negated subsection (-0.4 ppt.) while the MTL model performs better on the non-negated subsection (+0.3 ppt.).  
%WHAT CAN WE SAY ABOUT THIS CONFLICTING RESULTS ON BINARY AND NEGATED?
%
% Edit; see below, does this make sense? (erikve)
More detailed analysis would be needed to explain why the binary-task STL model outperforms the MTL model on our silver-standard negated sample. However, when it comes to the better performance of the MTL model on the non-negated sample -- for both the binary and fine-grained task --  one possible explanation is that learning about negation also enables the model to make more reliable predictions about sentiment bearing words that it has perhaps only seen in a negated context during training, but outside of negation during testing.

\begin{table}
\caption{Mean accuracy and standard devation of STL and MTL models on the negated and non-negated subsets of the SST test data.}

\footnotesize
\begin{minipage}{\textwidth}
\begin{tabular}{llll}
\hline\hline
Dataset & Model & \multicolumn{2}{c}{Sub-set accuracy} \\
\hline
& & negated & non-negated \\
\cmidrule(r){3-3}\cmidrule(r){4-4}
\multirow{2}{*}{SST-binary} & STL & 78.9 \sd{(1.2)} & 85.8 \sd{(0.9)}\\
							& MTL & \textbf{81.1 \sd{(1.5)}} & \textbf{87.1 \sd{(0.2)}}\\
\cmidrule(r){1-1}\cmidrule(r){2-2}\cmidrule(r){3-3}\cmidrule(r){4-4}
\multirow{2}{*}{SST-fine} & STL &\textbf{41.1 \sd{(1.4)}} & 47.6 \sd{(0.4)}\\
							& MTL & 40.7 \sd{(1.2)} & \textbf{47.9 \sd{(0.9)}}\\
\hline\hline
\end{tabular}
\end{minipage}
\label{table:analysis}
\end{table}

\subsection{Impact of data size}
\label{sec:negsize}
\begin{figure}
    \centering
    \includegraphics[width=\textwidth]{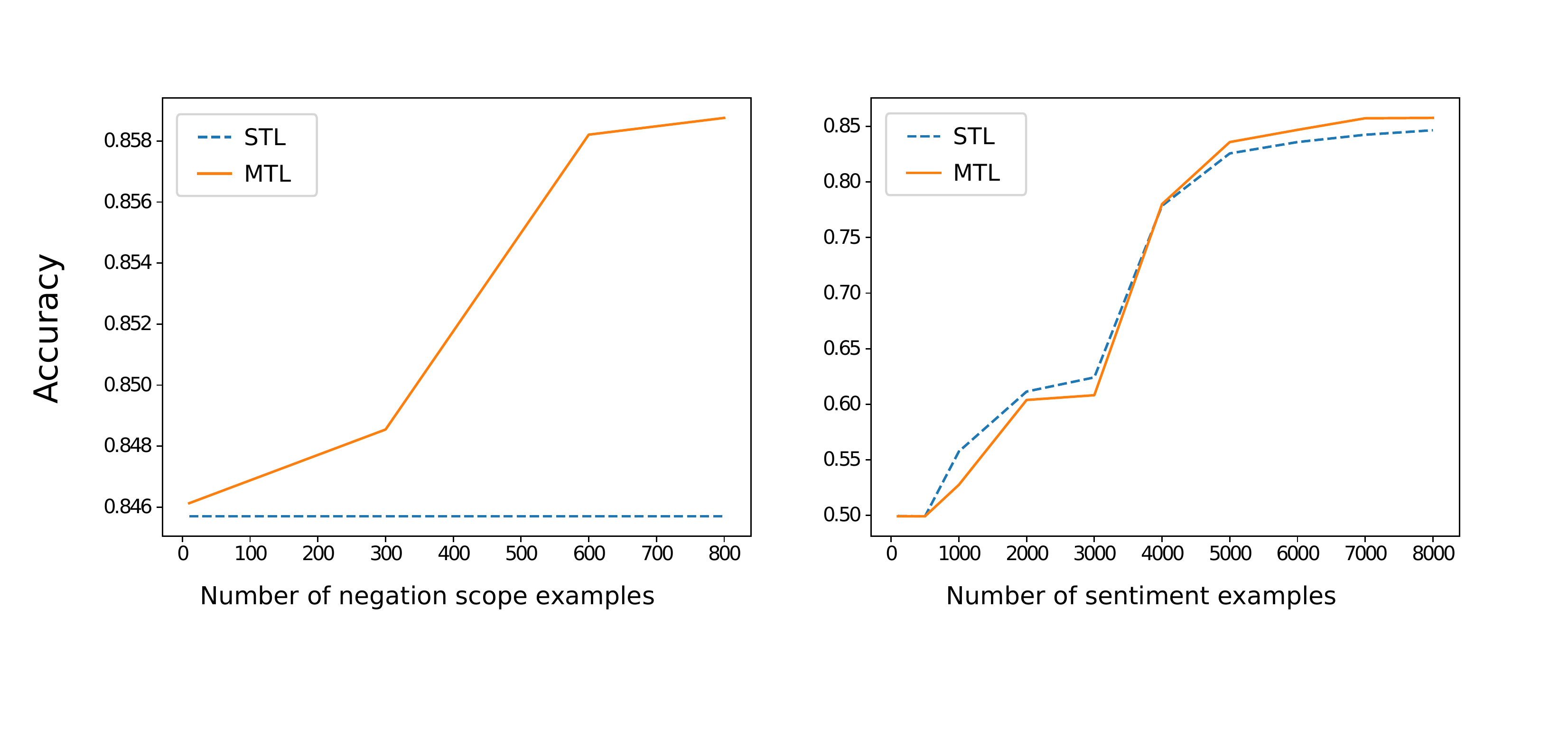}
    \caption{Mean accuracy on the SST-binary task when training MTL negation model with differing amounts of negation data from the SFU dataset (left) and sentiment data (right).}
    \label{figure:data_ablation}
\end{figure}

In order to better understand the effects of multi-task learning of negation detection, we 
%perform a data ablation study  
compute learning curves 
with respect to the negation data for 
the SST-binary setup. The model is given access to an increasing number of negation examples from the SFU dataset (from 10 to 800 in intervals of 100) and accuracy is calculated for each number of examples. Figure~\ref{figure:data_ablation} (left) shows that the MTL model improves over the baseline with as few as ten negation examples and plateaus somewhere near 600. An analysis on the SST-fine setup showed a similar pattern. 
% fixme: i started rambling a bit about learning curves; see if you think it makes sense to include: 
There is nearly always an effect of diminishing returns when it comes to adding training examples, but if we were to instead plot this learning curve with a log-scale on the x-axis, i.e. doubling the amount data for each increment, it would seem to indicate that having more data could indeed still prove useful, as long as there were sufficient amounts. In any case, regardless of the amount of data, exposing the model to a larger \textit{variety} of negation examples could also prove beneficial -- we follow up on this point in the next subsection. 

%\begin{figure}
%\includegraphics[width=\textwidth]{data_ablation.pdf}
%\caption[]{Mean accuracy on the SST-binary task when training MTL negation model with differing amounts of negation data from the SFU dataset.}
%\label{figure:data_ablation}
%\end{figure}

%\subsection{How does the amount of sentiment data affect MTL?}

While the previous experiment shows that models already improve with as few as 10 auxiliary examples, here we investigate whether a sentiment model benefits from multi-task learning more when there is limited sentiment data, as previous research has shown for other tasks \cite{Ruder2019}. We keep the amount of auxiliary training data steady, and instead vary the sentiment training data from 100--8000 examples.

%\begin{figure}
%\includegraphics[width=\textwidth]{sent_data_binary.pdf}
%\caption[]{Results of training with differing amounts of sentiment data.}
%\label{figure:sentdata}
%\end{figure}

Figure~\ref{figure:data_ablation} (right) shows that the performance of the STL model begins to plateau at around 5000 training examples (although note the comment about diminishing returns above). Until this point the MTL model performs either worse or similarly. From 5000 on, however, the MTL model is always better. Therefore, negation detection cannot be used to supplement a model when there is lacking sentiment data, but rather can improve a strong model. This may be a result of using a relevant auxiliary task which has a different labeling unit (sequence labeling vs. sentence classification), as other research \cite{Ruder2019} suggests that for similar tasks, we should see improvements with less main task data.

\subsection{Can we combine negation data despite differences in annotation?}

The previous experiment suggests that more negation data will not necessarily lead to large improvements. However, the model trained on the SFU negation dataset performs better on the SST-binary task, while the CD negation model is better on SST-fine. In this section, we ask whether a combination of the two negation data sets will give better results, despite the fact that they have conflicting annotations, see Section \ref{sec:data} above. 

We train an MTL model on the concatenation of the SFU and CD train sets (MTL-Combined) using the same hyperparameters as in the previous experiments. The results in Table \ref{table:combined} show that MTL-Combined performs worse than the MTL-SFU model on SST-binary, while it is the best performing model by a small margin ($p > 0.01$ with approximate randomization tests as described in Section \ref{sec:results}) on SST-fine. This shows that simply combining the negation data sets does not necessarily lead to improved MTL results, which is most likely due to the differences in the annotation schemes. 

\begin{table}
\caption{Combining the SFU and CD negation data (MTL-Combined) does not lead to large improvements.}
\begin{minipage}{\textwidth}
\begin{tabular}{lllll}
\hline\hline
Setup & STL &  MTL-SFU  & MTL-CD & MTL-Combined \\
\hline
SST-binary & 84.57 \sd{(1.0)} & \textbf{86.04 \sd{(0.3)}} & 85.43 \sd{(0.9)} & 85.54 \sd{(0.7)} \\
SST-fine & 46.49 \sd{(0.7)} & 46.75 \sd{(0.8)} & 47.33 \sd{(0.6)} & \textbf{47.42 \sd{(0.3)}} \\
\hline\hline
\end{tabular}
\end{minipage}
\label{table:combined}
\end{table}

\subsection{Scopes or cues}

As described in the initial sections, negation is usually represented by a negation cue and its scope. One interesting question to address is whether both of these are equally important for downstream use in our multi-task setup.
Here, we investigate whether it is enough to learn to identify only cues or only scopes. We train the MTL model from Section \ref{sec:model} to predict only cues or only scopes, and compare their results with the STL model and the original MTL model which predicts both.

\begin{table}
\caption{Mean accuracy and standard deviation on the sentiment task for the STL model, MTL models trained to predict only negation cues or only negation scope, and finally the MTL model trained to predict both scopes and cues.}

\footnotesize
\begin{minipage}{\textwidth}
\begin{tabular}{lllll}
\hline\hline
Setup & STL &  MTL-Cues  & MTL-Scopes & MTL-Both \\
\hline
SST-binary & 84.57 \sd{(1.0)} & 85.94 \sd{(0.2)} & 85.60 \sd{(1.3)} & \textbf{86.04 \sd{(0.3)}} \\
SST-fine & 46.49 \sd{(0.7)}	& 46.43 \sd{(1.1)}	&	46.48 \sd{(1.2)}	&	 \textbf{47.33 \sd{(0.6)}}\\
\hline\hline
\end{tabular}
\end{minipage}
\label{table:cue_scope}
\end{table}

The results of each experiment are shown in Table \ref{table:cue_scope}. Learning to predict only cues or only scopes performs worse than the MTL model trained to predict both. Additionally, learning to predict only one of the two elements also performs worse than STL on the fine-grained setup. This indicates that it is necessary to learn to predict both scopes and cues. One likely explanation for this is that the cue predictions in turn benefits scope predictions. On the SST-binary, the MTL-Cues model performs better than the MTL-Scopes model, while the opposite is true for the SST-fine task, indicating that it is more important to correctly predict the negation scope for the fine-grained setting than the binary.

\subsection{Can models trained on phrase-level data improve with MTL?}
\label{sec:phrase-level}

Besides the sentence-level annotations we have used so far, the Stanford Sentiment Treebank also contains sentiment annotations at each node of a constituent tree (i.e., its constituent phrases) for all sentences (statistics are shown in Table \ref{table:SST-phrase}).  Although originally intended to enable recursive approaches to sentiment, it has also been shown that training a non-recursive model with these annotated phrases leads to models that are better at capturing compositionality effects \cite{Iyyer2015}. This is likely because, given a sentence such as ``The movie was not great'', models are explicitly shown that ``great'' is positive while ``not great'' is negative. A relevant question is therefore whether this phrase-level annotation of sentiment reduces the need for explicit negation annotation. In this section, we compare training on these sentiment-annotated phrases to multi-task learning on negation annotations, and also the combination of these two approaches.

\begin{table}
\caption{Number of training, development, and test examples for the Stanford Sentiment Treebank phrase-level data.}

\footnotesize
\begin{minipage}{\textwidth}
\begin{tabular}{lrrrr}
\hline\hline
Setup & \multicolumn{1}{c}{Train} & \multicolumn{1}{c}{Dev.} & \multicolumn{1}{c}{Test} & \multicolumn{1}{c}{Total} \\
\hline
SST-binary & 75,646 & 10,099 & 19,972 & 105,717 \\
SST-fine & 155,019 & 20,173 & 40,195 & 215,387 \\
\hline\hline
\end{tabular}
\end{minipage}
\label{table:SST-phrase}
\end{table}

We train the STL model from Section~\ref{sec:model} on the phrase-level SST data (STL-phrase) and compare with the MTL model trained with phrase-level SST data and with negation as an auxiliary task (MTL-phrase). In order to fairly compare with models trained only on sentence-level annotation, we test on the sentence-level SST data described in Section~\ref{sec:data}. The results in Table~\ref{table:phraselevel} show that even though the largest gains are found by training on the phrase-level data, multi-task learning of negation still provides small but consistent gains. This indicates that while end-to-end models may learn some compositional functions implicitly when trained on phrase-level data, there is still room for further improvements by combining this with training on explicit negation annotations in addition. However, while the MTL approach is generally applicable to any sentiment dataset, the phrase-level annotations are particular to the SST data. 

\begin{table}
\caption{Mean accuracy and standard deviation of STL, MTL, and STL-Phrase-level models.}

\footnotesize
\begin{minipage}{\textwidth}
\begin{tabular}{lllll}
\hline\hline
Setup & STL & MTL & STL-Phrase & MTL-Phrase \\
\hline
SST-binary & 84.57 \sd{(1.0)} & 86.04 \sd{(0.3)} & 87.82 \sd{(0.4)} & \textbf{88.18 \sd{(0.5)}}\\
SST-fine & 46.49 \sd{(0.7)} & 47.33 \sd{(0.6)} & 49.10 \sd{(1.2)} & \textbf{49.71 \sd{(0.4)}}\\
\hline\hline
\end{tabular}
\end{minipage}
\label{table:phraselevel}
\end{table}

\subsection{Evaluating the negation component: a case for transfer learning?}
\label{sec:transfer}

Although our interest in \textit{negation} modeling in this paper is primarily tied to its influence on sentiment analysis, we do, however, also want to evaluate negation performance in isolation, just to make sure the model is reasonable. There are a number of evaluation metrics used for negation detection. For example, \F scores can be computed with respect to cues or scopes or both, either requiring an exact match of predicted spans or  allowing for partial matches, or evaluating on the token-level. A range of different metrics were implemented for the *SEM 2012 shared task on negation, see \cite{Morante2012} for an overview. In this section, we report \F for cues and scopes separately, both on the token-level. The latter corresponds to the measure called `scope tokens' in \cite{Morante2012} and \cite{Fancellu2016}. 

%Following the policy of the *SEM scorer, punctuation tokens are ignored in evaluation \cite{Morante2012}. 
% <-- fixme: consider updating our scorer to align with this for the final version?

Table~\ref{table:negresults} compares our best performing MTL sentiment models with a set-up where the negation component of the architecture -- corresponding to only the first-layer BiLSTM+CRF as shown in Figure~\ref{figure:model} -- is trained as a single-task model for negation prediction. The single-task negation model achieves a  token-level scope \F score of 89.23 on the SFU data and 75.38 on the CD data, while the MTL model reaches 74.69 and 63.81, respectively. As we are optimizing the MTL models for sentiment, the single-task models achieve much better token-level scope \F scores (14.5 ppt. on the SFU data, 11.6 on CD). For comparison, the best performing system \cite{ReaVelOvr12b} on CD with respect to the same metric in the *SEM 2012 shared task \cite{Morante2012} achieved an \F of 85.26. 

An analysis of the common errors shows that neither the STL nor MTL models generalize well to morphological negation cues, \eg ``\underline{un}likely'', that have not been seen in training. This is not surprising, given that neither model has access to subtoken information. Of course, this also affects the scopes, as the models rely on  predicted cues. Additionally, the MTL model has difficulty identifying long scopes. 

\begin{table}
\caption{Token-level \F for the SFU and ConanDoyle-neg (CD) negation tasks. The single task negation models outperform the multi-task (MTL) models. Note that the MTL models are not tuned to optimize negation, this being the auxiliary task.}
\begin{minipage}{\textwidth}
\footnotesize
\begin{tabular}{llll}
\hline\hline
Setup & Evaluation & SFU & CD\\
\hline
\multirow{2}{*}{STL}& Scopes & 89.23 \sd{(0.4)} & 75.38 \sd{(0.9)}\\
    & Cues   & 97.84 \sd{(1.0)} & 86.25 \sd{(0.2)}           \\
\multirow{2}{*}{MTL} & Scopes & 80.45 \sd{(7.5)} & 63.81 \sd{(2.8)}\\
    & Cues   & 94.84 \sd{(2.2)} & 84.14 \sd{(1.8)} \\
\hline\hline
%\bottomrule
\end{tabular}
\end{minipage}
\label{table:negresults}
\end{table}

Given that the lower BiLSTM(+CRF) component does not achieve strong results for the auxiliary negation task when trained in the multi-task setup, it is logical to ask if better sentiment predictions can be obtained by starting from a better performing negation model. To test this, we explore a \textit{transfer learning} approach where we pre-train the negation component with a single-task negation objective as described above. 

In contrast to the MTL set-up, with transfer learning we first optimize the negation parameters and afterwards use these parameters to initialize the lower BiLSTM layer of the sentiment model (cf. Figure~\ref{figure:model}). These pre-trained parameters are then further fine-tuned when train the overall model on sentiment data, but using a reduced learning rate. Note that this continued training is no longer multi-task learning, however, as the entire network is only supervised by the sentiment task.

\begin{table}
\caption{Mean accuracy and standard deviation of single-task, multi-task, and transfer-learning models.}

\footnotesize
\begin{minipage}{\textwidth}
\begin{tabular}{llll}
\hline\hline
Setup & STL & MTL & Transfer \\
\hline
SST-binary & 84.57 \sd{\sd{(1.0)}} & \textbf{86.04 \sd{(0.3)}} & 85.57 \sd{(0.7)} \\
SST-fine & 46.49 \sd{(0.7)} & \textbf{47.33 \sd{(0.6)}} & 47.17 \sd{(0.4)} \\
\hline\hline
\end{tabular}
\end{minipage}
\label{table:transferlearning}
\end{table}

Table~\ref{table:transferlearning} shows that while transfer learning based on initializing the sentiment model with a pre-trained negation model does show improvements over the single-task sentiment model (1 ppt. on binary and 0.7 on fine-grained), it performs worse than multi-task learning. Counterintuitively, having a better negation detection model, in terms of performance on the negation data sets, does not lead to better results on the sentiment main task. %HOW CAN IT BE EXPLAINED?

We also consider that the poor performance of the transfer learning approach may be due to overfitting to the training data, which only contains negated examples. However, further experiments training with balanced data (50\% negated and 50\% non-negated) give poorer performance overall (85.3 \F for both STL and MTL scope-level), indicating that this is not the root of the problem.

%For future work we would like to follow up on these experiments by trying to combine transfer and multi-task learning to see whether this could yield improvements. 

\subsection{Comparing negation detection to other common auxiliary tasks}
\label{sec:otherauxtasks}

In multi-task learning for natural language processing it is common to employ a number of auxiliary tasks, which range from simple tasks, (predicting word frequency), to morphosyntactic tasks (chunking, dependency relation classification), to semantic tasks (semantic frame detection, super-sense tagging). In this section, we compare common auxiliary tasks and their effect on sentiment analysis. Specifically, we train the MTL model from Section \ref{sec:model} on three additional auxiliary tasks: POS tagging, multi-word detection (MWE) (identifying multi-word expressions, \ie \textit{by the way}, \textit{cope with}), and super-sense tagging (SEM) (assigning course-grained semantic types to verbs and nouns). %Table \ref{table:auxiliarytasksexample} shows an example sentence which has been annotated for these three auxiliary tasks. 

The data for the auxiliary tasks comes from the STREUSLE dataset \cite{Schneider2015}, which contains sentences from the \textsc{Review} section of the English Web Treebank \cite{Bies2012}, which have been enriched with multi-word and super-sense annotations. Table \ref{table:auxdata} shows the statistics of the data sets, as well as the entropy and kurtosis of the labels. Here entropy indicates the amount of uncertainty in the label distribution, while kurtosis indicates the skewness. These measures have been shown to correlate well to the usefulness of auxiliary tasks in previous work \cite{Alonso2017,Bingel2017}.

%\begin{table}
%\caption{An example of a sentence from the STREUSLE dataset annotated for Part-of-Speech (POS), multi-word detection (MWE), and super-sense tagging (SEM).}
%\footnotesize
%\begin{minipage}{\textwidth}
%\begin{tabular}{l|llllllllll}
%\hline\hline
%Tokens &We &  enjoyed & our & stay & at & the & Vintage & Hostel & . \\
%POS & PRP  & VBD & PRP\$ & NN & IN & DT & NNP & NNP & . \\
%MWE & -&-  & -& - & -& -& 1:1 & 1:2 & -\\
%SEM & - & v.emotion & - & n.event &  - & - & n.group & - & - \\
%\hline\hline
%%\bottomrule
%\end{tabular}
%\end{minipage}
%\label{table:auxiliarytasksexample}
%\end{table}

\begin{table}
\caption{Train and test splits available for each auxiliary task, as well as label entropy and kurtosis.}
\footnotesize
\begin{minipage}{\textwidth}
\begin{tabular}{lrrrr}
\hline\hline
Dataset &\multicolumn{1}{c}{\# Train} & \multicolumn{1}{c}{\# Test} & \multicolumn{1}{c}{Label Entropy} & \multicolumn{1}{c}{Label Kurtosis} \\
\hline
SFU & 800 & 96 & 0.21 & 0.2 \\
ConanDoyle-neg & 842 & 235 & 1.02 & $-$0.8 \\
POS Tagging & 2,451 & 272 & 3.04 & 2.8 \\
Multi-word Detection & 2,451 & 272 & 0.64 & 29.8\\ 
Super-sense Tagging & 2,451 & 272 & 2.16 & 82.7\\
\hline\hline
%\bottomrule
\end{tabular}
\end{minipage}
\label{table:auxdata}
\end{table}

\begin{table}
\caption{Accuracy on SST-fine with POS tagging, multi-word detection (MWE), and super-sense tagging (SEM) as auxiliary tasks.}
\footnotesize
\begin{minipage}{\textwidth}
\begin{tabular}{llllll}
\hline\hline
Dataset & STL & Negation & POS & MWE & SEM \\
\hline
SST-fine & 46.49 \sd{(0.7)} &  \textbf{47.33 \sd{(0.6)}} & 46.44 \sd{(0.9)} & 46.93 \sd{(0.6)} & 46.11 \sd{(0.5)} \\
\hline\hline
%\bottomrule
\end{tabular}
\end{minipage}
\label{table:auxtasks}
\end{table}

The results are shown in Table \ref{table:auxtasks}. POS tagging and super-sense tagging perform worse than the baseline single-task model, while multi-word detection and negation detection show improvements. The MTL negation model, however, is still the best performing model, which demonstrates the importance of negation on sentiment classification. The fact that multi-word detection is helpful may correlate to the importance of multi-word idioms in expressions of sentiment \cite{Williams2015,liu-etal-2017-idiom,jochim-etal-2018-slide,barnes-etal-2019}.

Both the SFU and CD data sets have low label kurtosis, but also have relatively low label entropy. This partially aligns with previous research \cite{Alonso2017,Bingel2017}, which suggests that for an auxiliary task to improve the main task, the entropy of the labels should be high (implying the task should not be trivial to learn), and the kurtosis should be low (the labels should not have an overly long-tailed distribution). The fact that the multi-word task is more helpful, however, seems to indicate that the appropriateness of the auxiliary task for the main task is more important than the specific dataset properties. %\cite{Bjerva2017} suggests conditional entropy and mutual information to be better measures, but these require jointly labeled data, which we do not have.

% RESULTS WITH TRAINING ON BOTH NEGATED AND NON-NEGATED EXAMPLES
% MTL - 85.32 (0.8)
% Transfer - 85.29 (0.9)

\section{Conclusion and future work}
\label{sec:conclusion}

This paper introduces a multi-task learning approach to incorporating explicitly annotated negation information into a sentiment classifier. We employ a cascading architecture where one BiLSTM is shared between the sentiment and negation tasks and feeds into a higher-level BiLSTM dedicated only to sentiment prediction (also using skip connections). We show that using negation as an auxiliary task helps improve the main task of sentiment analysis and that the effect persists across several different standard data sets. While we only report results for English here, for future work we plan to extend the experiments to other languages that have annotations for both tasks available, \eg Spanish.

 The extensive analysis of the results reveals several effects of using negation detection as an auxiliary task. On the one hand, we find that even a small amount of annotated negation data allows a multi-task learner to improve, while on the other hand, it is necessary to have enough sentiment data to achieve relatively good performance in order to see improvements. We further show that detection of both negation cues and scopes as an auxiliary task is preferable over detecting only one of these.

In this work, negation cues were always modeled on the token-level, but morphological negation is another important realization of negation that our current model does not fully take into account. Adding character-level information to the network could be of interest in the future. Moreover, it may also be useful separate the cue and scope classification, in order to improve the negation module.

We have noted several places that the two data sets for negation employed in this work operate with slightly different annotation schemes. Due to the fact that these data sets are taken from different domains and genres, it has not been possible for us to compare the effect of these differing annotation choices systematically. Another avenue for future work would therefore be to compare the effect of different annotation schemes for negation by comparing the use of the ConanDoyle-neg and the re-annotated version of this dataset dubbed NegPar \cite{Liu:Fan:Web:18} in our multi-task setup for sentiment analysis. 

Regarding multi-task learning, we demonstrate that it is possible to use an auxiliary task with different labeling units (token-level sequence-labeling) to improve the main task (sentence-level classification). Additionally, we show that negation detection is a more suitable auxiliary task for sentiment analysis than other standard auxiliary tasks, such as POS tagging, multi-word detection, or super-sense tagging. Finally, our experiments on transfer learning indicate that multi-task learning may provide a better framework to leverage negation information, but other approaches to transfer learning, such as freeze-thaw \cite{felbo-etal-2017-using} or discriminative fine-tuning \cite{howard-ruder-2018-universal} may give better results. We also want explore the combination of multi-task learning and transfer learning (i.e. continued training of pre-trained negation layers in an MTL set-up).

Although we only experiment with negation in this work, there are many other linguistic and paralinguistic phenomena, \ie speculation, multi-word expressions, sarcasm, etc., which also affect sentiment classification \cite{Cruz2016,FARIAS2017113,barnes-etal-2019}. Here we have shown that explicit training via hierarchical multi-task learning is a viable way to incorporate some of this information. In the future, we would like to incorporate other sources of linguistic knowledge in a similar fashion.

\label{lastpage}

\bibliography{lit}

\end{document}